\documentclass[11pt]{article}

\usepackage{acl}
\usepackage{times}
\usepackage{latexsym}
\usepackage[T1]{fontenc}
\usepackage[utf8]{inputenc}
\usepackage{microtype}
\usepackage{amsmath, amssymb}
\usepackage{booktabs}
\usepackage{multirow}
\usepackage{graphicx}
\usepackage{enumitem}
\usepackage{placeins}

\newcommand{\Tok}{\ensuremath{T_{\mathrm{ok}}}}
\newcommand{\Twrong}{\ensuremath{T_{\mathrm{wrong}}}}
\newcommand{\Trepair}{\ensuremath{T_{\mathrm{repair}}}}
\newcommand{\BASM}{\textsc{BASM}}

\title{When Not to Imitate: Boundary-Aware Skill Memory for Reliable Tool-Use LLM Agents}
\newcommand{\equalmark}{\textsuperscript{*}}
\newcommand{\corrmark}{\textsuperscript{\ensuremath{\dagger}}}

\author{
\textbf{Zihan Lin}\textsuperscript{1,2,3}\equalmark
\quad
\textbf{Zhenyu Chen}\textsuperscript{3,4}\equalmark
\quad
\textbf{Jiawen Wei}\textsuperscript{1}\equalmark
\quad
\textbf{Xiaohan Wang}\textsuperscript{1}\corrmark
\quad
\textbf{Jie Cao}\textsuperscript{3} \\
\quad
\textbf{Jiajun Chai}\textsuperscript{1}
\quad
\textbf{Wei Lin}\textsuperscript{1}
\quad
\textbf{Guojun Yin}\textsuperscript{1}
\quad
\textbf{Ran He}\textsuperscript{2,3}\corrmark
\\[3pt]
\textsuperscript{1}
Meituan
\\
\textsuperscript{2}
School of Advanced Interdisciplinary Sciences,
University of Chinese Academy of Sciences
\\
\textsuperscript{3}
MAIS\&NLPR, Institute of Automation, Chinese Academy of Sciences
\\
\textsuperscript{4}
Zhongguancun Academy
\\
[3pt]
\small
\equalmark Equal contribution.
\quad
\corrmark Corresponding authors.
\\
\small
\texttt{wangxiaohan17@meituan.com}
\quad
\texttt{ran.he@ia.ac.cn}
}

\begin{document}
\maketitle

 \begin{abstract}
Extracting skills from past successes is critical for the efficient evolution of Large Language Model (LLM) agents. Prevailing agent self-evolution paradigms typically rely on a core assumption: equipping LLMs with skill memories derived from successful trajectories will monotonically improve their problem-solving capabilities. However, probe analyses reveal that extracting skills solely from successful trajectories traps the model in a \textbf{Skill Imitation Trap}. For tasks that resemble past successes but require different tools, retrieving more skills paradoxically increases the model's confidence in wrong tool calls---procedure skills raise the wrong-tool margin by $47\%$ over a memory-free baseline. To overcome this limitation, we propose \textbf{Boundary-Aware Skill Memory} (BASM), which augments each skill with explicit boundary fields---applicability conditions, risk cues, avoidance rules, and recovery notes. These fields transform each retrieved skill from an unconditional action template into state-conditioned guidance: the agent applies the skill when its conditions hold, suppresses inapplicable tool calls when they do not, and issues targeted repairs when execution fails. Across three agent benchmarks and four model scales, BASM consistently outperforms success-distilled skill-memory baselines: it improves task success rate by up to $23.8\%$ on AppWorld, accuracy by up to $5.0\%$ on BFCL, and reduces attack success rate by $4.6\%$ on AgentDojo, while simultaneously reducing average AppWorld steps by up to $6.6\%$ relative to the memory-free baseline.
\end{abstract}

\section{Introduction}
\label{sec:intro}

Large language model (LLM) agents are increasingly deployed in complex, interactive environments that demand multi-step tool use, API orchestration, and long-horizon planning~\citep{yao2022react,lin2026resrl,xi2025rise,lin2026rest,lu2026contextual,chen2025toolforge}. A central challenge in building such agents is enabling them to improve over time without constant human supervision. Skill acquisition, which enables agents to extract, store, and reuse procedural knowledge from past experience, has emerged as a critical mechanism for agent self-evolution~\citep{wang2023voyager,zhu2023ghost,zhao2024expel,lin2025awpo}. By distilling successful trajectories into reusable skill memories, agents can avoid redundant exploration and transfer solutions across structurally similar tasks~\citep{wang2026implicithierarchicalgrpodecoupling,mi2026skill}.

Existing approaches have explored how skills are represented, acquired, retrieved, and evolved across a range of settings. Early agents such as {Voyager} and {ExpeL} compressed successful trajectories into reusable programs or verbal lessons~\citep{wang2023voyager,zhao2024expel}; recent work has extended this to online skill induction, RL-driven distillation, and cross-agent library transfer~\citep{wang2025inducing,tang2025agent,lu2026skill0,wang2026skillx,wang2026self}. Despite their varied perspectives, these approaches largely treat success-distilled skills as positive transfer: once retrieved as relevant, a skill is presumed useful, while its validity {boundary} remains implicit.   

This omission matters when semantic relevance diverges from decision validity: a retrieved skill appears relevant but lacks boundary cues for whether its action pattern still applies. By probing model behavior at varying retrieval depths, we identify a failure mode we term the \textbf{Skill Imitation Trap} (Figure~\ref{fig:retrieval-trap}): the model's preference for a semantically similar but inapplicable tool increases as more success-distilled skills are retrieved, causing skill memory to underperform a memory-free baseline in inapplicable states. Mechanistically, attention analysis pinpoints the cause: when a retrieved skill matches the current query, success-distilled memory shifts the model's attention toward procedural content rather than boundary cues, treating the retrieved skill as an unconditional action template. This teaches the model to focus on {how} a past task was solved, with no mechanism to assess {whether} the present state actually meets the conditions for that solution.

\begin{figure*}[t]
    \centering
    \includegraphics[width=0.98\textwidth]{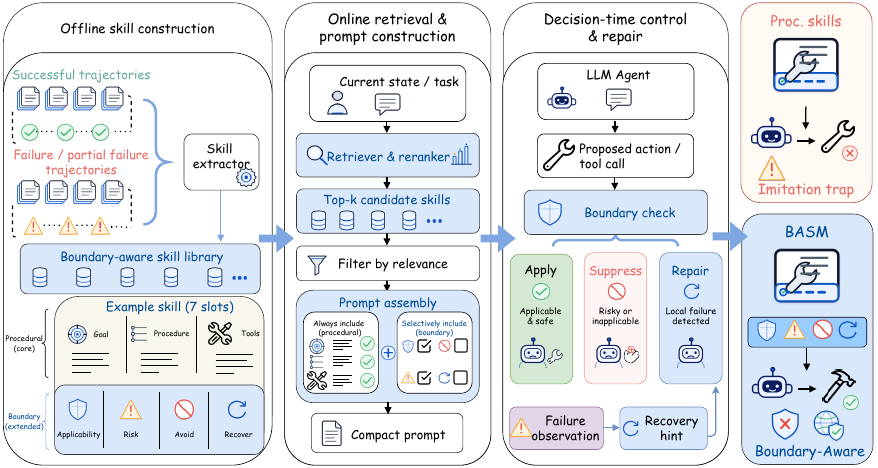}
    \caption{\textbf{Overview of \BASM{}.}
\BASM{} builds a boundary-aware skill library from logged trajectories, storing each skill as a procedural core---goal, procedure, and tools---with boundary fields covering applicability, risk, avoidance, and recovery.
At inference, \BASM{} retrieves candidate skills, selectively exposes boundary evidence under a fixed token budget, and checks proposed tool calls to decide whether to apply, suppress, or repair.
This transforms skill reuse from blind imitation into validity-scoped transfer, suppressing wrong-tool preference while preserving efficient reuse and local recovery.}
    \label{fig:basm-overview}
        \vspace{-0.5em}
\end{figure*}

To overcome the Skill Imitation Trap, we propose \textbf{Boundary-Aware Skill Memory} (BASM), a skill-memory method that augments procedural knowledge with explicit boundary conditions.
Figure~\ref{fig:basm-overview} illustrates how BASM transports boundary evidence from offline extraction to online retrieval and decision-time checking.
Unlike success-distilled skill memory, BASM treats a retrieved skill as state-contingent evidence rather than an unconditional action template, enabling the agent to apply the skill when its assumptions hold, suppress risky tool calls when they do not, and repair local failures when execution deviates. Probe analyses confirm that boundary fields function as a state-conditioned attention modulator: the model consults boundary content more in risky and repair states than in safe ones. Moreover, same-prompt knockout shows that blocking boundary-span attention recovers $69.4\%$ of the imitation-trap effect, confirming that boundary fields are a causal anti-imitation channel. Across BFCL, AppWorld, and AgentDojo on four model scales, BASM consistently outperforms success-distilled skill memory baselines: it improves task success rate by up to $23.8\%$ on AppWorld and accuracy by up to $5.0\%$ on BFCL; it reduces attack success rate by $4.6\%$ on AgentDojo while reducing average AppWorld steps by up to $6.6\%$ relative to the memory-free baseline; and it achieves Pareto improvements in both task utility and safety on Qwen3-14B.

Our contributions are as follows:
\begin{enumerate}[leftmargin=*]

\item \textbf{Skill Imitation Trap probe analyses.}
We identify and characterize the Skill Imitation Trap through mechanism-level probe analyses. Using wrong-tool margin measurements, attention attribution, and same-prompt knockout, we show that success-distilled skill memory amplifies wrong-tool confidence in risky states (wrong-tool margin increases by $47\%$ relative to memory-free baseline) and routes its imitation effect through procedural attention at decision tokens. Critically, ablating boundary-span attention recovers $69.4\%$ of the imitation effect, confirming that boundary fields serve as a causal anti-imitation channel.

\item \textbf{Boundary-Aware Skill Memory design.}
We propose BASM, a seven-slot boundary-aware skill schema---goal, procedure, tools, applicability conditions, risk cues, avoidance rules, and recovery notes---together with a budgeted retrieval-and-formatting pipeline, a boundary checker for prompt-side enforcement, and a runtime repair module for stateful environments. This design transforms retrieved skills from unconditional action templates into boundary-aware records that guide when a skill should be applied, suppressed, or locally repaired.

\item \textbf{Broad experimental validation.}
We evaluate BASM across three complementary benchmarks (BFCL, AppWorld, AgentDojo) and four model scales spanning $8$B to $397$B parameters. BASM consistently improves over success-distilled skill-memory baselines: AppWorld success rate improves by up to $23.8\%$; BFCL accuracy improves by up to $5.0\%$; AgentDojo attack success rate drops by $4.6\%$; average AppWorld steps fall by up to $6.6\%$ relative to the memory-free baseline; and Qwen3-14B achieves Pareto improvements in both utility and safety.
Attention, logit, and knockout probes provide a coherent mechanistic account of these gains.

\end{enumerate}

\section{Related Work}
\label{sec:related}
\paragraph{Self-Evolution Agents.}
LLM tool use has evolved from early demonstrations---the Reason-Act-Observe loop of \citep{yao2022react}, self-supervised API invocation in Toolformer~\citep{schick2023toolformer}, and large-scale API mastery in ToolLLM~\citep{qin2024toolllm}---to a central paradigm in agent research, yet these approaches treat each task independently without accumulating reusable knowledge. Skill memory emerged to address this limitation. \citep{wang2023voyager} and \citep{zhu2023ghost} pioneered skill libraries that store successful behaviors as executable code or text-based knowledge for compositional reuse. \citep{zhao2024expel} and \citep{wang2024awm} distilled experience into natural-language insights and reusable workflows without parameter updates. \citep{shinn2023reflexion} pursued episode-level verbal reflection rather than a persistent skill library. More recent work has extended the paradigm along orthogonal axes: program-based online induction~\citep{wang2025inducing}, hierarchical skill knowledge bases~\citep{wang2026skillx}, RL-driven skill internalization~\citep{lu2026skill0}, and cross-framework memory sharing~\citep{tang2025agent}. Despite their varied perspectives, all these approaches share a common premise: skills distilled from successful experience, when retrieved as relevant, should benefit the current decision. This positive-transfer assumption leaves the {boundary} of a skill under-specified, and can break down in practice---recent work has noted that retrieved skills sometimes cause models to over-imitate procedures rather than adapt them to novel tasks~\citep{wang2026skillx}, an empirical signal consistent with the Skill Imitation Trap we identify through mechanism-level analysis in Section~\ref{sec:probe}.

\paragraph{Skill Retrieval and Applicability.}
BASM's retrieval module builds on RAG~\citep{lewis2020retrieval,jeong2024adaptive} and demonstration retrieval for in-context learning~\citep{rubin2022learning,lu2022fantastically}. Standard retrieval scores measure relevance, not applicability: neither retrieved documents nor procedurally similar skills carry an explicit signal for when retrieved content should be {suppressed}~\citep{yoran2023making,asai2024self}. BASM addresses this by augmenting each skill with boundary fields and incorporating a boundary relevance score into reranking and prompt formatting, so that retrieved skills function as validity-scoped records rather than unconditional action templates.
The idea of conditioning skill execution on explicit preconditions has a precedent in reinforcement learning: \citep{konidaris2014constructing} formalized skill applicability via initiation sets within the options framework~\citep{sutton1999between}. BASM transplants this boundary-conditioning idea to LLM agents, replacing hand-crafted symbolic predicates with natural-language fields extracted from agent trajectories and enforced at retrieval, formatting, and execution time.

\section{Probe Analysis}
\label{sec:probe}

We use probe analysis to establish two claims: success-distilled skill memory can increase wrong-tool preference in inapplicable states, and BASM counteracts this effect through boundary-aware attention at the decision token. The analysis first diagnoses how retrieval depth affects wrong-tool preference, then shows how boundary evidence is read, how it suppresses wrong-tool logits, and how attention knockout tests the causal role of the boundary path.

\begin{figure}[h]
\centering
\includegraphics[width=0.9\linewidth]{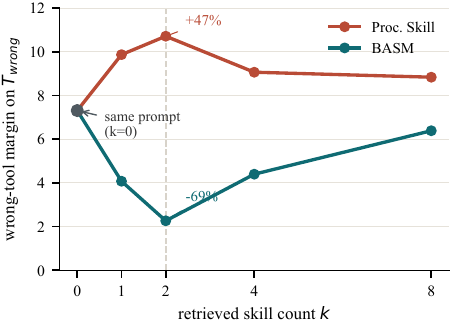}
\caption{Success-distilled skill memory (\textbf{Proc.\ Skill}) amplifies wrong-tool preference with retrieval depth; \BASM{} reverses this trend. }
\label{fig:retrieval-trap}
    \vspace{-0.5em}
\end{figure}

\subsection{Diagnostic Setup}
\label{sec:probe-setup}

Let \(x_t\) denote the prompt prefix immediately before the agent generates its next action, and let \(d_t\) be the first generated structural token of that action. In BFCL, \(d_t\) is the first token of the function name; in stateful environments, it is the first content token of the next code, tool, or action step. We partition decision points into three task-state buckets: \Tok{} (the retrieved skill matches the correct tool and should be applied), \Twrong{} (a semantically similar skill is retrieved but is inapplicable and should be suppressed), and \Trepair{} (a local failure has occurred and the agent should repair rather than repeat).

The central diagnostic is the {wrong-tool margin}. Let \(\mathcal{Y}_{\mathrm{wrong}}\) be the first tokens of semantically similar but incorrect function names, and let \(\mathcal{Y}_{\mathrm{ctrl}}\) contain the first tokens of the correct function and abstention tokens. We define
\begin{equation}
m_{\mathrm{wrong}}(x_t)
= \max_{y \in \mathcal{Y}_{\mathrm{wrong}}}\ell_t(y)
- \max_{y \in \mathcal{Y}_{\mathrm{ctrl}}}\ell_t(y).
\end{equation}
A positive value means that the model locally prefers an incorrect tool over the correct or abstention alternatives. We measure this margin at \(d_t\), before environment feedback can intervene. This logit-level diagnostic exposes the model's immediate tool preference, even when later repair steps might correct the final action.

\subsection{Retrieval Trap}
\label{sec:probe-negative-transfer}

Figure~\ref{fig:retrieval-trap} shows that retrieving more success-distilled skills can amplify, rather than correct, risky tool choices. At \(k=0\), both memory configurations produce the same prompt, so their wrong-tool margins coincide at \(7.31\). Once retrieved skills are added, the curves diverge: success-distilled skill memory rises to \(10.71\) at \(k=2\), whereas BASM falls to \(2.26\). This reversal shows that additional procedural evidence can strengthen preference for a semantically similar but invalid tool, challenging the monotonicity assumption behind success-distilled skill memory~\citep{wang2023voyager,zhao2024expel}. The design implication is direct: a skill should encode not only {how} to execute a procedure, but also when that procedure should not be used.

\begin{figure*}[t]
\centering
\includegraphics[width=\linewidth]{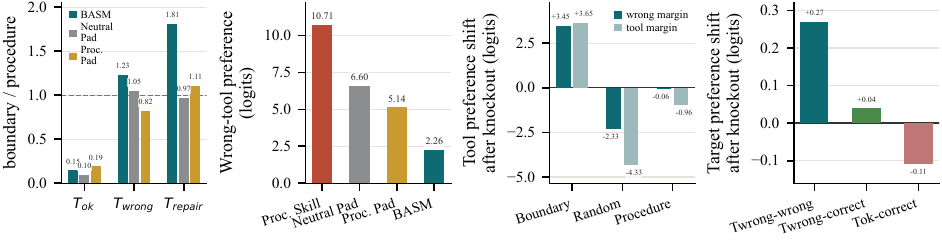}
\caption{\textbf{Boundary fields suppress wrong-tool preference through a state-conditioned attention mechanism.} BASM reduces the wrong-tool margin from $10.71$ to $2.26$, a suppression that padding controls cannot explain. Knocking out boundary-span attention restores the imitation-trap direction. $T_\mathrm{ok}$: applicable state; $T_\mathrm{wrong}$: inapplicable state; $T_\mathrm{repair}$: post-failure state}
\label{fig:probe-summary}
    \vspace{-0.5em}
\end{figure*}

\subsection{Mechanism Analysis}
\label{sec:probe-mechanism}

\paragraph{Boundary attention.}
We first test whether the model reads boundary fields at the decision point. For a retrieved skill, let \(P\) be the token span of its procedure field and \(B\) be the union of its boundary fields: applicability conditions, risk cues, avoidance rules, and recovery notes. For attention layer \(l\) and head \(h\), we compute the length-normalized attention from the decision token to a span \(X\):
\begin{equation}
\begin{aligned}
\alpha_{X}^{(l,h)}(x_t)
&= \frac{1}{|X|}\sum_{i \in X} A_{d_t,i}^{(l,h)}, \\
\eta^{(l,h)}(x_t)
&= \frac{\alpha_{B}^{(l,h)}(x_t)}{\alpha_{P}^{(l,h)}(x_t)+\epsilon}.
\end{aligned}
\end{equation}
The ratio \(\eta\) measures how much the decision token attends to boundary evidence relative to procedural content. The first panel of Figure~\ref{fig:probe-summary} shows a state-dependent pattern: under BASM, the ratio is low in \Tok{} (\(0.15\)), rises in \Twrong{} (\(1.23\)), and peaks in \Trepair{} (\(1.81\)). Same-length controls with neutral padding or extra procedure text do not reproduce this pattern. Thus, the shift is not explained by prompt length or formatting; it reflects the model's use of boundary evidence when the retrieved procedure becomes risky or requires repair.

\paragraph{Wrong-tool suppression.}
The second panel of Figure~\ref{fig:probe-summary} links this attention pattern to the action distribution. Its y-axis reports wrong-tool preference in logits: larger values mean stronger preference for a semantically similar but invalid tool. Proc.\ Skill yields a high wrong-tool preference of \(10.71\) on \Twrong{} examples. Same-length controls reduce it to \(6.60\) and \(5.14\), showing that additional context can partially soften the action distribution. BASM lowers the value further to \(2.26\), well below both controls. This gap shows that boundary evidence provides a suppression signal beyond length and formatting, allowing the model to reject invalid procedural imitation while retaining the retrieved skill as contextual evidence.

\subsection{Attention Knockout}
\label{sec:probe-knockout}

The third and fourth panels of Figure~\ref{fig:probe-summary} test whether boundary attention is causally responsible for this suppression. We keep the prompt tokens fixed and ablate only the attention from the decision token \(d_t\) to the boundary span \(B\). All bars report the signed shift after knockout, \(\Delta m = m_{\mathrm{knockout}} - m_{\mathrm{BASM}}\), so positive values indicate that removing an attention path increases the corresponding target preference. On Qwen3-8B \Twrong{} examples, boundary knockout raises wrong-tool preference by \(+3.45\) logits and the broader tool preference by \(+3.65\) logits. Random same-length and procedure-span knockouts do not produce the same shift, showing that the effect is specific to boundary access rather than a generic attention perturbation. The fourth panel repeats the diagnostic on Qwen3-14B: \Twrong{}--wrong measures invalid-tool preference in inapplicable states, \Twrong{}--correct measures correct-tool preference in the same states, and \Tok{}--correct measures correct-tool preference when the retrieved skill is applicable. Boundary knockout selectively increases \Twrong{}--wrong (\(+0.27\)), leaves \Twrong{}--correct nearly unchanged (\(+0.04\)), and slightly decreases \Tok{}--correct (\(-0.11\)). This pattern shows that BASM is not imposing a general penalty on tool use; instead, boundary attention specifically suppresses invalid imitation while largely preserving preference for correct actions.
Extended evidence is reported in Appendix~\ref{app:probe-figures}.

\begin{table}[h]
\centering
\small
\caption{\textbf{BFCL results on Qwen3-32B and Qwen3.5-397B-A17B.} We report accuracy across four multi-turn subsets. Best in \textbf{bolded} and second are \underline{underlined}.}
\resizebox{\linewidth}{!}{
\begin{tabular}{lccccc}
\toprule
\multicolumn{6}{c}{\textbf{Qwen3-32B}} \\
\midrule
Method & Base & Long Context & Miss Func & Miss Param & Overall \\
\midrule
Base & 52.0 & 41.5 & 45.5 & 35.0 & 43.50 \\
Proc. Skill & \textbf{60.5} & 41.5 & 49.0 & 36.0 & \underline{46.75} \\
App. Skill & \underline{60.0} & 41.5 & 47.0 & \textbf{37.0} & 46.38 \\
BASM & 57.5 & \textbf{42.5} & \textbf{51.0} & \underline{36.5} & \textbf{46.88} \\
\midrule
\multicolumn{6}{c}{\textbf{Qwen3.5-397B-A17B}} \\
\midrule
Base & 62.0 & 51.5 & 50.0 & \textbf{44.5} & 52.00 \\
Proc. Skill & \textbf{64.0} & \textbf{56.5} & \underline{53.0} & 40.0 & \underline{53.38} \\
App. Skill & \underline{63.0} & 55.0 & 50.0 & 38.5 & 51.63 \\
BASM & \textbf{64.0} & \underline{55.5} & 51.5 & \underline{44.0} & \textbf{53.75} \\
\bottomrule
\end{tabular}}
    \vspace{-1.0em}
\label{tab:qwen32-bfcl}
\end{table}

\begin{table}[h]
\centering
\small
\caption{\textbf{AppWorld results on Qwen3-32B and Qwen3.5-397B-A17B.} BASM achieves the highest AppWorld task success rate among the tested methods.}
\resizebox{\linewidth}{!}{
\begin{tabular}{llcc}
\toprule
Model & Method & Success & Avg. Steps \\
\midrule
\multirow{4}{*}{Qwen3-32B}
& Base & 62.50 & 12.46 \\
& Proc. Skill & 66.07 & \textbf{11.61} \\
& App. Skill & 60.12 & 12.03 \\
& BASM & \textbf{76.19} & 11.71 \\
\midrule
\multirow{4}{*}{Qwen3.5-397B-A17B}
& Base & 35.12 & 13.50 \\
& Proc. Skill & 52.98 & 12.64 \\
& App. Skill & 52.98 & \textbf{12.61} \\
& BASM & \textbf{54.76} & \textbf{12.61} \\
\bottomrule
\end{tabular}}
\label{tab:qwen32-appworld}
\end{table}

\section{Method}
\label{sec:method}

BASM is a boundary-aware skill-memory pipeline for tool-use agents. Given an offline set of logged trajectories, it constructs a boundary-aware skill library, retrieves relevant skills for a new decision state, formats them under a fixed context budget, and optionally performs local repair in stateful environments.

\subsection{Skill Schema}
\label{sec:method-overview}

Formally, let \(\mathcal{D}\) be a skill library and let \(\rho_{\mathcal{D}}\) be a retrieval-and-formatting operator that maps a task state \(x_t\) to an in-context memory block. A skill-conditioned agent samples the next action from
\begin{equation}
a_t \sim \pi_\theta(\cdot \mid x_t, \rho_{\mathcal{D}}(x_t)).
\end{equation}
Proc.\ Skill improves the procedural information in \(\rho_{\mathcal{D}}\), but does not constrain when the retrieved procedure should be applied. BASM changes the library object and the formatting operator so that the prompt exposes an explicit boundary signal alongside procedural content.
The boundary signal is realized through a structured seven-slot skill schema. BASM represents each reusable behavior as a seven-slot boundary-aware skill:
\begin{equation}
s = (g,\, \pi,\, \mathcal{T},\, \mathcal{A},\, \mathcal{R},\, \mathcal{N},\, \mathcal{F}),
\end{equation}
where \(g\) is the skill goal, \(\pi\) is the procedural specification, \(\mathcal{T}\) is the tool set, \(\mathcal{A}\) denotes applicability conditions, \(\mathcal{R}\) denotes risk cues, \(\mathcal{N}\) denotes avoidance rules, and \(\mathcal{F}\) denotes recovery notes. The first three fields constitute the procedural component, while the last four collectively form the boundary component $B(s)$. We define the boundary text of a skill as
\begin{equation}
B(s) = \mathcal{A}(s) \oplus \mathcal{R}(s) \oplus \mathcal{N}(s) \oplus \mathcal{F}(s),
\end{equation}
where \(\oplus\) denotes ordered text composition. This decomposition reflects the distinct functional roles of the two components: \(\pi\) enables the agent to execute a known procedural pattern, whereas $B(s)$ provides the contextual evidence necessary to determine whether that pattern is applicable.

Each boundary slot serves a distinct role: applicability conditions specify when the procedure is safe to invoke; risk cues identify observations that render a semantically similar skill unsafe; avoidance rules enumerate actions or parameters to suppress under particular states; and recovery notes prescribe one-step corrections after local failures. Together, these fields give each retrieved skill an explicit validity scope, which the pipeline enforces at retrieval, formatting, and execution time (\S\ref{sec:pipeline}).

\subsection{Method Design}
\label{sec:pipeline}

\paragraph{Skill library construction.}
BASM constructs the skill library offline from logged trajectories. Let \(\mathcal{K}^{+}\) be the set of successful trajectories and \(\mathcal{K}^{-}\) be failed or partially failed trajectories, and let \(\mathcal{U}\) be the set of all extraction anchors derived from these trajectories. Extraction is anchored at two granularities: step-level extraction derives functional skills from plan steps, while tool-level extraction derives atomic skills from concrete tool invocations and their associated observations. For an anchor \(u\), the extractor receives a successful context \(\tau_u^{+}\) and, when available, a nearby failure context \(\tau_u^{-}\), then emits a candidate skill
\begin{equation}
\hat{s}_u = f_{\mathrm{ext}}(u,\, \tau_u^{+},\, \tau_u^{-}).
\end{equation}
The successful context supplies \(g,\pi,\mathcal{T}\); the failure context is used only as boundary evidence for \(\mathcal{R},\mathcal{N},\mathcal{F}\). 

The raw extractor output is passed through a deterministic validation operator \(\psi\). This operator enforces strict JSON parsing, required fields, list-valued boundary slots, non-empty risk cues and avoidance rules, and tool-schema validity. Candidate skills that fail validation after retry are discarded, ensuring that library construction degrades gracefully rather than halting on malformed extractions. The resulting library is
\begin{equation}
\mathcal{D}_{\mathrm{BASM}}
= \bigl\{\,\psi(\hat{s}_u) \mid u \in \mathcal{U},\;
\psi(\hat{s}_u)\neq \varnothing\,\bigr\}.
\end{equation}
This precision-first construction prioritizes boundary specificity over recall, ensuring that every retained skill carries actionable boundary evidence rather than generic procedural summaries.

\paragraph{Retrieval and prompt formatting.}
At inference time, BASM retrieves skills with a multi-view score that combines semantic relevance, tool overlap, and step/goal overlap. For decision state \(x_t\) and skill \(s\), let \(e(\cdot)\) be a text embedding function, \(v_s\) be the concatenation of the goal and procedure fields of \(s\), \(\mathcal{T}_{x_t}\) be the set of tools referenced in the current state, and \(J(\cdot,\cdot)\) be Jaccard overlap over tool names. The retrieval score is
\begin{equation}
\begin{aligned}
\mathrm{score}(x_t,s)
={} & \lambda_{\mathrm{sem}}\cos\bigl(e(x_t),e(v_s)\bigr) \\
&+ \lambda_{\mathrm{tool}}\, J(\mathcal{T}_{x_t},\mathcal{T}_{s}) \\
&+ \lambda_{\mathrm{step}}\,\mathrm{overlap}(x_t,g_s,\pi_s),
\end{aligned}
\end{equation}
where $g_s$ and $\pi_s$ denote the goal and procedure fields of skill $s$, and $\lambda_{\mathrm{sem}}, \lambda_{\mathrm{tool}}, \lambda_{\mathrm{step}}$ are weighting coefficients. The top-$k$ retrieved set is
\begin{equation}
\mathcal{S}_k(x_t)
= \operatorname{TopK}_{s\in\mathcal{D}_{\mathrm{BASM}}}
\mathrm{score}(x_t,s).
\end{equation}

The prompt formatter operates under a fixed token budget, motivated by the probe finding (Figure~\ref{fig:retrieval-trap}) that indiscriminate retrieval amplifies rather than mitigates wrong-tool preference. For each retrieved skill, the procedural component \(g,\pi,\mathcal{T}\) is always exposed. Let $r_B(x_t, s)$ denote the boundary relevance score measuring the overlap between the current state $x_t$ and the boundary fields of skill $s$. Boundary fields are exposed only when $r_B(x_t,s)$ exceeds a threshold $\gamma$. The formatted memory block is
\begin{equation}
\begin{aligned}
\rho_{\mathcal{D}}(x_t)
={} & \mathrm{Proc}(\mathcal{S}_k) 
\oplus \mathrm{Compress}_{\beta}\!\Bigl(\\
&\bigl\{B(s) \mid s\in\mathcal{S}_k,\,
r_B(x_t,s)>\gamma\bigr\}\Bigr),
\end{aligned}
\end{equation}
where \(\beta\) is the maximum boundary-token budget.

\paragraph{Boundary-conditioned execution.}
BASM separates prompt-side guidance from runtime control. The boundary checker compares the proposed tool call $a_t$ against retrieved boundary fields, blocking or revising calls that violate applicability, match an avoidance rule, or repeat a known failed pattern. Let $\mathcal{S}_k^{*}(x_t,a_t)$ denote the subset of retrieved skills whose tools or goals match $a_t$.
\begin{equation}
\begin{aligned}
G(x_t,a_t,\mathcal{S}_k)
= \prod_{s\in\mathcal{S}_k^{*}(x_t,a_t)}
\mathbb{I}\bigl[&
a_t \not\models \mathcal{N}(s) \\
&\;\wedge\;
x_t \models \mathcal{A}(s)
\bigr],
\end{aligned}
\end{equation}
where $\mathbb{I}[\cdot]$ is the indicator function. In practice, $G$ is evaluated via deterministic string/schema checks first, with LLM-assisted judgment reserved for ambiguous natural-language conditions.

\paragraph{Local failure repair.}
Recovery notes $\mathcal{F}$ deliver their primary benefit through a runtime loop in stateful environments. In single-turn settings, $\mathcal{F}$ steers tool selection at inference time; no execution loop exists in which a repair
action can be issued. In multi-turn stateful settings, $\mathcal{F}$ additionally drives a repair loop: when observation $o_{t+1}$ signals a local failure, the repair module retrieves a targeted hint
\begin{equation}
  h_t = \mathrm{Repair}(x_t,\,a_t,\,o_{t+1},\,\mathcal{S}_k),
\end{equation}
and injects it into the next decision step in place of full skill re-retrieval.

\begin{table}[h]
\centering
\small
\caption{\textbf{Results on Qwen3 variants.} BASM turns skill memory from procedural reuse into controlled transfer.}
\resizebox{0.5\textwidth}{!}{
\begin{tabular}{lcccc}
\toprule
Benchmark & Base & Proc. Skill & App. Skill & BASM \\
\midrule
\multicolumn{5}{c}{\textbf{Qwen3-8B}} \\
\midrule
BFCL Acc. ($\uparrow$)          & 34.13 & 33.88 & \underline{35.25} & \textbf{38.88} \\
AppWorld SR ($\uparrow$)        & 29.76 & \underline{33.33} & \underline{33.33} & \textbf{36.31} \\
AgentDojo Utility ($\uparrow$)  & \textbf{40.99} & 31.51 & 31.61 & \underline{35.93} \\
AgentDojo ASR ($\downarrow$)    & \underline{8.01} & 10.01 & 10.54 & \textbf{5.90} \\
\midrule
\multicolumn{5}{c}{\textbf{Qwen3-14B}} \\
\midrule
BFCL Acc. ($\uparrow$)          & 39.88 & 37.38 & \underline{41.00} & \textbf{42.50} \\
AppWorld SR ($\uparrow$)        & 33.33 & 46.43 & \underline{53.57} & \textbf{57.14} \\
AgentDojo Utility ($\uparrow$)  & 29.93 & 31.40 & \underline{31.61} & \textbf{33.51} \\
AgentDojo ASR ($\downarrow$)    & \underline{3.79} & 6.01 & 6.11 & \textbf{2.95} \\
\bottomrule
\end{tabular}}
\label{tab:main}

\end{table}

\begin{table}[h]
\centering
\small
\caption{\textbf{Component ablation on Qwen3-8B.} \textit{Base}, \textit{Proc. Skill}, \textit{App. Skill}, and \textit{BASM} follow Table~\ref{tab:main}. \textit{Prompt-BASM} uses BASM-style boundary prompting without the runtime repair loop; \textit{Repair-Only} uses the runtime repair loop without boundary-enriched prompts; \textit{BASM $-$ Avoid} removes avoidance rules; \textit{BASM $-$ Recovery} removes recovery notes.}
\resizebox{0.5\textwidth}{!}{
\begin{tabular}{lcccc}
\toprule
Mode & BFCL ($\uparrow$) & AppWorld ($\uparrow$) & Utility ($\uparrow$) & ASR ($\downarrow$) \\
\midrule
Base             & 34.13 & 29.76 & \textbf{40.99} & 8.01 \\
Proc. Skill           & 33.88 & 33.33 & 31.51 & 10.01 \\
App. Skill           & 35.25 & 33.33 & 31.61 & 10.54 \\
\midrule
Prompt-BASM       & \underline{38.38} & 29.17 & 31.93 & 10.54 \\
Repair-Only       & 34.13 & 25.60 & 30.87 & 7.17 \\
\textbf{BASM}      & \textbf{38.88} & \textbf{36.31} & \underline{35.93} & \textbf{5.90} \\
BASM $-$ Avoid   & 38.00 & \textbf{36.31} & 30.87 & 6.95 \\
BASM $-$ Recovery & 37.00 & \underline{33.93} & 31.82 & \underline{6.85} \\
\bottomrule
\end{tabular}}
\label{tab:ablation}
    \vspace{-1.0em}
\end{table}

\section{Experiments}
\label{sec:experiments}

\subsection{Setup}
\label{sec:setup}

\paragraph{Benchmarks.}
We evaluate on three benchmarks covering complementary capability tiers. \textbf{BFCL v3}~\citep{patil2025berkeley} tests multi-turn function calling (800 instances across \textit{base}, \textit{long\_context}, \textit{miss\_func}, and \textit{miss\_param} subsets); the miss\_func and miss\_param subsets directly instantiate the Skill Imitation Trap by requiring the agent to reject semantically similar but inapplicable tool calls. \textbf{AppWorld}~\citep{trivedi2024appworld} evaluates stateful multi-step task completion via Python-style API calls in a persistent app environment (168 \textit{test\_normal} tasks); it is the primary testbed for BASM's runtime repair loop. \textbf{AgentDojo v1.2.2}~\citep{debenedetti2024agentdojo} measures the utility--safety tradeoff under \textit{tool\_knowledge} prompt-injection attacks across four task suites, reporting Utility ($\uparrow$) and Attack Success Rate ($\downarrow$).

\paragraph{Models and baselines.}
We evaluate \textbf{Qwen3-8B}, \textbf{Qwen3-14B}, \textbf{Qwen3-32B}, and \textbf{Qwen3.5-397B-A17B}~\citep{yang2025qwen3} as the agent backbones. All models are used with thinking mode disabled; Qwen3's extended chain-of-thought reasoning in thinking mode would conflate skill-boundary effects with multi-step internal deliberation, making controlled comparison unreliable. We set temperature to $0.001$. All configurations share an identical retrieval substrate, backbone family, and evaluation pipeline, enabling direct comparison across capacity tiers.

We compare four memory configurations.
\textbf{Base} is the memory-free agent.
\textbf{Proc.\ Skill} retrieves the goal, procedure, and tools of successful past trajectories, instantiating the procedural-memory view of skills as reusable execution knowledge distilled from experience~\citep{wang2023voyager,zhao2024expel,wang2025inducing,wang2026skillx}.
\textbf{App.\ Skill} augments Proc.\ Skill with applicability conditions only, isolating the classical idea that a reusable skill should specify when it can be applied, as in action preconditions and option initiation sets~\citep{konidaris2014constructing}.
\textbf{BASM} is the full boundary-aware schema with applicability conditions, risk cues, avoidance rules, and recovery notes.

\subsection{Main Results}
\label{sec:main-results}

Table~\ref{tab:main} supports three aggregate conclusions.
First, the Skill Imitation Trap is visible at benchmark scale: Proc.\ Skill underperforms Base on BFCL for both Qwen3-8B and Qwen3-14B, confirming that procedural memory alone can amplify wrong-tool preference in inapplicable states.
Second, BASM consistently overcomes this trap, outperforming the strongest skill-memory baselines on BFCL and AppWorld across both model sizes.
Third, AgentDojo reveals a capacity-dependent safety--utility pattern: on Qwen3-8B, BASM achieves the largest ASR reduction among memory methods while incurring only a minor utility cost; on Qwen3-14B, it achieves Pareto improvements, yielding both the highest utility and the lowest ASR among all memory configurations.

Tables~\ref{tab:qwen32-bfcl} and~\ref{tab:qwen32-appworld} extend these results to larger models. On Qwen3-32B, BASM achieves the best aggregate BFCL accuracy, with its largest margin on Miss Func (the subset most directly instantiating the Skill Imitation Trap); on AppWorld it reaches $76.19\%$ success, improving over Proc.\ Skill ($66.07\%$), while reducing average steps by $6.0\%$ relative to Base ($12.46$ to $11.71$). On Qwen3.5-397B-A17B, the ranking is preserved: BASM reaches $53.75$ on BFCL versus $53.38$ for Proc.\ Skill, and $54.76\%$ AppWorld success versus $52.98\%$, with average steps reduced by $6.6\%$ relative to Base ($13.50$ to $12.61$).
These lower average-step counts relative to Base indicate that BASM's gains do not require longer interaction trajectories: boundary fields make retrieved skills more targeted, and the repair module supports local recovery when execution deviates---consistent with the mechanism established in Section~\ref{sec:probe}.

\subsection{Ablation Study}
\label{sec:ablation}
{Prompt-BASM} nearly matches BASM on BFCL, so boundary-conditioned prompting explains most of the function-calling gain; {Repair-Only} stays at Base on BFCL ($34.13\%$) and drops below it on AppWorld ($25.60\%$ vs.\ $29.76\%$), showing that runtime repair without boundary-conditioned prompts provides no benefit on single-step tasks and can actively disrupt multi-step execution without proper applicability guidance. On AppWorld, however, the ranking flips: Prompt-BASM falls below Base ($29.17\%$ vs. $29.76\%$), while BASM reaches $36.31\%$. The two components are therefore complementary: boundary prompting sharpens tool-choice precision, and runtime repair carries benefit in stateful multi-step recovery.

Removing {avoidance rules} lowers BFCL by $0.88$ points and leaves AppWorld unchanged, but it also reduces AgentDojo utility from $35.93$ to $30.87$ and raises ASR from $5.90$ to $6.95$. This indicates that avoidance rules contribute not only to function-selection precision but also to safety-sensitive suppression in adversarial tool-use settings. Removing {recovery notes} has a broader performance effect: BFCL falls by $1.88$ points and AppWorld by $2.38$ points; AgentDojo utility decreases to $31.82$, and ASR rises to $6.85$. Recovery notes therefore influence prompt-side tool-choice logits and stateful recovery behavior, consistent with the probe findings in Section~\ref{sec:probe-mechanism}. In contrast, adding applicability conditions alone (App. Skill) improves BFCL over Proc. Skill by $1.37$ points but raises ASR by $0.53$ points, suggesting that partial boundary specification can inadvertently increase susceptibility to injected instructions that satisfy applicability criteria on the surface.

\section{Conclusion}
\label{sec:conclusion}
This work reveals a failure mode in skill-memory agents that goes beyond procedural knowledge gaps: without explicit validity boundaries, retrieved skills can amplify wrong-tool preference rather than suppress it.
BASM closes this gap by equipping each skill with explicit boundary fields---covering applicability, risk, avoidance, and recovery---enforced at retrieval, prompt formatting, and execution to transform skill reuse from unconditional imitation into state-conditioned decision-making.
Evaluated across BFCL, AppWorld, and AgentDojo at model scales from $8$B to $397$B, BASM consistently outperforms success-distilled skill-memory baselines: task success improves by up to $23.8\%$ on AppWorld; function-calling accuracy by up to $5.0\%$ on BFCL; attack success rate drops by $4.6\%$ on AgentDojo; and average AppWorld steps decrease by up to $6.6\%$ relative to the memory-free baseline.
Knockout probes causally attribute wrong-tool suppression to boundary-span attention, not to prompt length or added context; ablations confirm that boundary-conditioned prompting and runtime repair provide complementary benefits---showing that the key to reliable skill reuse is not more procedures, but explicit validity boundaries around each one.

\newpage
\bibliography{custom}
\newpage

\section*{Limitations}

This work focuses on text-based tool-use agents in benchmarked environments where task goals, observations, and tool feedback can be expressed in natural language. 
While this setting covers a broad class of practical agent tasks, future work could study boundary-aware skill memory in richer interactive settings, such as multimodal agents, dynamically changing tool APIs, or long-running workflows with delayed feedback. 
In addition, our current implementation represents boundary conditions as explicit natural-language fields, which makes the mechanism interpretable and easy to inspect; an interesting direction is to explore more compact or adaptive boundary representations that can be selected under tighter context budgets. 
Finally, our evaluation emphasizes autonomous tool-use performance, safety under adversarial instructions, and interaction efficiency; extending the analysis to human-agent collaboration or personalized tool preferences may further clarify how boundary-aware memories should be adapted across users and deployment settings.
As with any method that improves autonomous tool-use agents, broader deployment may raise risks if agents are connected to high-impact tools without appropriate permissioning, audit logging, or human approval, although BASM itself is designed to make tool use more state-aware and easier to inspect.
We use publicly available benchmarks and environments under their respective licenses and terms of use, and any released code, prompts, or evaluation scripts associated with this work will include explicit license information; third-party artifacts will not be redistributed beyond what their original licenses permit.

\section*{Artifact Use}
We use existing benchmarks and environments, including AppWorld, BFCL, and AgentDojo, only for their intended research and evaluation purposes and in accordance with their documented access conditions. 
Artifacts produced by this work, such as code, prompts, evaluation scripts, and derived skill memories, are intended to support research on tool-use agents; any derived artifacts based on benchmark trajectories should remain within research contexts and should not be redistributed or deployed in ways that exceed the original artifacts' licenses or terms of use.

We do not collect real user data in this work; our experiments use publicly available benchmark environments and model-generated trajectories for research evaluation. 
Before releasing any derived artifacts, such as prompts, trajectories, or skill memories, we screen them for personally identifying information, credentials, API keys, unique user identifiers, and offensive content using automated pattern checks and manual inspection of representative samples. 
Any detected sensitive fields are removed, anonymized, or excluded from release, and benchmark-derived artifacts are shared only when consistent with the original access conditions.
We use the official evaluation partitions provided by each benchmark and do not introduce new train/dev/test splits or a new human-collected dataset; the reported statistics therefore correspond to the benchmark instances used for evaluation.
We use the official evaluation scripts for BFCL v3, AppWorld, and AgentDojo v1.2.2 with their default evaluator settings unless otherwise stated. 
We do not use additional NLP preprocessing, normalization, or text-overlap packages such as NLTK, SpaCy, or ROUGE; model choices, decoding parameters, maximum generation length, evaluated subsets, and metrics are reported in Section~\ref{sec:experiments}.
This work does not involve human-subject data collection, participant recruitment, paid annotation, user studies, or intervention with human participants. 
Therefore, participant-facing instructions, recruitment and payment procedures, data-consent procedures, and ethics review board approval are not applicable. 
AI assistants were used for language editing, polishing, formatting suggestions, and drafting support for some non-technical explanatory text. 
All research ideas, method design, experiments, results, analysis, and final manuscript content were reviewed, verified, and approved by the authors; AI assistants were not used to generate experimental results or replace author judgment.

\appendix

\section{Additional Experiments}
\label{app:probe-figures}

This appendix expands the probe evidence behind the {Skill Imitation Trap}. The main text shows the central reversal: success-distilled memory does not simply provide helpful experience; under semantically similar but inapplicable states, it becomes an imitation prior that pulls the model toward the retrieved procedure. BASM changes the role of memory. Instead of storing a bare success template, it stores a procedure together with its boundary of validity. The probes below make this mechanism visible at four levels: decision-token attention, next-action logits, same-prompt intervention, and realized call behavior. Together, they support a single interpretation: boundary fields act as a state-conditioned attention modulator that preserves useful procedural recall in \Tok{} while suppressing overconfident imitation in \Twrong{} and guiding recovery in \Trepair{}.

\subsection{Mechanism Replication and Logit Evidence}
\label{app:replication-logits}

\begin{figure}[h]
\centering
\includegraphics[width=\linewidth]{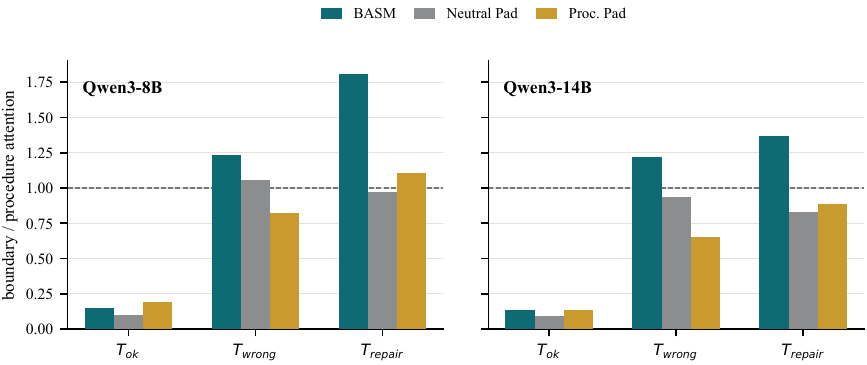}
\caption{\textbf{Boundary attention replication.} Length-normalized boundary/procedure attention across task-state buckets for Qwen3-8B and Qwen3-14B. Under BASM, decision-token attention shifts from the procedure toward boundary fields precisely in risky and repair states, while equal-length controls fail to reproduce this state-conditioned pattern.}
\label{fig:app-attention-ratio}
\end{figure}

\paragraph{Figure~\ref{fig:app-attention-ratio}: boundary fields are read when the state becomes risky.}
Figure~\ref{fig:app-attention-ratio} tests whether the added boundary slots are merely present in the prompt or are actually used at the decision point. The results indicate that, under BASM, the boundary/procedure attention ratio is low in \Tok{}, rises substantially in \Twrong{}, and rises again in \Trepair{}. This pattern is consistent with the intended boundary-aware behavior. In a safe state, the retrieved skill should behave like a normal reusable procedure, so attention can remain concentrated on the procedural recipe. In a risky state, however, the same procedure becomes dangerous because semantic similarity is no longer equivalent to applicability. The model therefore shifts attention toward applicability conditions, risk cues, avoidance rules, and recovery notes. This state-dependent shift is the core of boundary-aware memory: BASM does not make the agent uniformly conservative; it makes the agent consult the boundary exactly when imitation becomes unsafe.

The two equal-length controls sharpen this interpretation. Irrelevant padding and procedure-padding controls preserve much of the prompt length and schema position, but they do not reproduce the same risk-sensitive boundary/procedure profile. This rules out the simplest alternative explanation that the effect is caused by adding more tokens, slowing down the model, or changing the formatting template. The 14B curve follows the same qualitative pattern as the 8B curve, showing that the phenomenon is not an idiosyncrasy of a single backbone. The magnitude need not be identical across model sizes: larger models can distribute contextual evidence across more internal pathways. What matters for the mechanism is the invariant ordering---safe states read less boundary information, while risky and repair states read more.

\begin{figure}[h]
\centering
\includegraphics[width=\linewidth]{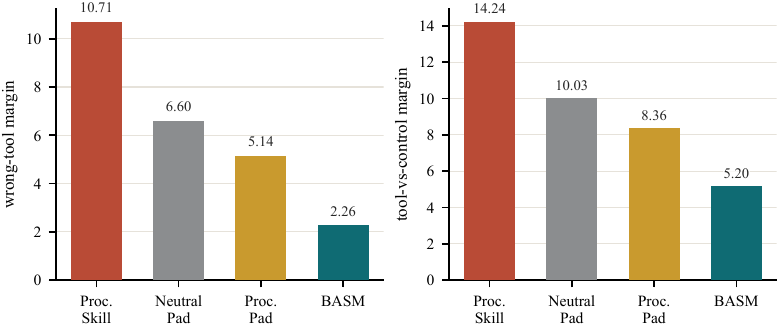}
\caption{\textbf{Detailed \Twrong{} logit margins.} BASM reduces both the specific wrong-tool margin and the broader tool-vs-control margin more than equal-length controls, showing that boundary semantics suppresses the risky action attractor rather than merely changing prompt length or formatting.}
\label{fig:app-logit-margins}
\end{figure}

\paragraph{Figure~\ref{fig:app-logit-margins}: boundary attention changes the action distribution.}
Figure~\ref{fig:app-logit-margins} connects the attention shift to the model's next-action preference. On \Twrong{}, Proc.\ Skill produces a high wrong-tool margin of $10.71$, meaning that the retrieved but currently invalid tool has become a strong local attractor. Equal-length controls partially reduce this margin ($6.60$ for irrelevant padding and $5.14$ for procedure padding), which is expected because extra context and schema regularity can mildly dilute a retrieved procedure. BASM, however, drives the margin down to $2.26$. The difference between $5.14$ and $2.26$ is the important part: once length and formatting are controlled, explicit boundary semantics still provides a much stronger suppression signal.

The right panel reaches the same conclusion with a broader tool-vs-control margin. BASM reduces this margin from $14.24$ under Proc.\ Skill to $5.20$, while the two controls remain substantially higher. Thus, BASM is not only discouraging one mislabeled function token; it is lowering the model's general tendency to commit to an inappropriate tool action in a state where the retrieved skill should be rejected. This directly supports the central claim: the failure mode is not lack of skill knowledge, but lack of a boundary around skill reuse.

\subsection{Same-Prompt Knockout Evidence}
\label{app:knockout-controls}

\begin{figure}[h]
\centering
\includegraphics[width=\linewidth]{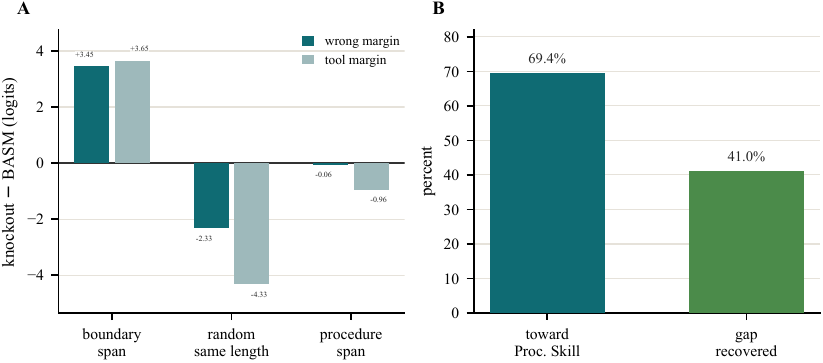}
\caption{\textbf{Boundary knockout controls.} Same-prompt attention knockout identifies the boundary span as the causal anti-imitation channel. Masking boundary attention pushes BASM back toward the Proc.\ Skill imitation pattern, recovering $69.4\%$ of the wrong-tool gap, while random same-length and procedure-span masks fail to restore the imitation pattern.}
\label{fig:app-knockout-8b}
\end{figure}

\paragraph{Figure~\ref{fig:app-knockout-8b}: removing boundary attention reopens the imitation trap.}
Figure~\ref{fig:app-knockout-8b} provides the strongest intervention evidence because the prompt tokens are kept fixed. We only mask attention from the decision token to the boundary span, leaving the rest of the memory block untouched. Under this same-prompt intervention, the wrong-tool margin rises by $+3.45$ logits, and the broader tool margin rises in the same direction. In other words, once the decision token can no longer read the boundary fields, the model moves back toward the success-distilled behavior that BASM was designed to prevent. The recovery analysis makes the effect more interpretable: boundary knockout recovers $69.4\%$ of the Proc. Skill gap for the wrong-tool margin and $41.0\%$ for the broader tool margin.

The controls are equally important. A random same-length mask does not produce the same effect, and masking procedure-span attention does not restore the imitation trap. This means the result is not a generic consequence of deleting context or damaging attention. The specific path from the decision token to the boundary span is what counteracts procedural imitation. The knockout therefore turns the boundary fields from a plausible prompt-engineering addition into a mechanistically active component: they are not merely ancillary prompt annotations, but an anti-imitation channel that keeps retrieved experience from becoming an unconditional action template.

\begin{figure}[h]
\centering
\includegraphics[width=\linewidth]{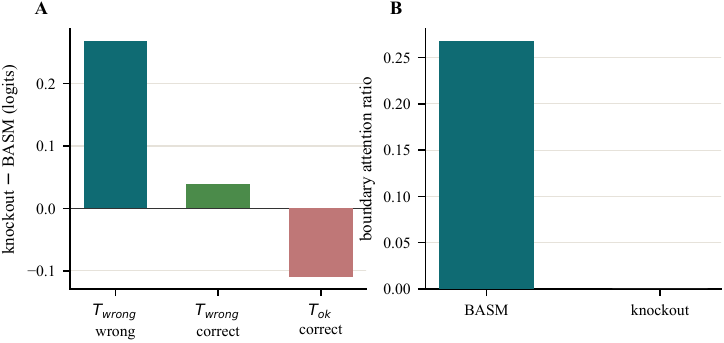}
\caption{\textbf{Qwen3-14B lightweight knockout.} Boundary knockout again moves the \Twrong{} wrong-tool margin upward while suppressing the boundary-attention path. The smaller magnitude is consistent with a more distributed implementation in the larger model, but the causal direction matches the 8B intervention.}
\label{fig:app-knockout-14b}
\end{figure}

\paragraph{Figure~\ref{fig:app-knockout-14b}: the larger model follows the same causal direction.}
Figure~\ref{fig:app-knockout-14b} repeats the intervention on Qwen3-14B. The effect is smaller than on Qwen3-8B but follows the same direction: boundary knockout increases the \Twrong{} wrong-tool margin, produces only a small change in the \Twrong{} correct-tool margin, and moves the \Tok{} correct-tool margin in the opposite direction. This selectivity is consistent with a boundary mechanism: the boundary path matters most when the retrieved procedure is tempting but invalid, rather than acting as a blanket penalty against tool use.

We interpret the smaller magnitude as a scale-dependent version of the same mechanism. A larger model may integrate boundary evidence through more distributed contextual routes, so blocking a single span-level path perturbs the final logits less strongly than in the 8B model. Even so, the intervention still moves the risky margin in the predicted direction and suppresses the boundary-attention path, confirming that the causal signature of boundary-aware retrieval survives a change in model scale.

\subsection{Where the Boundary Signal Appears}
\label{app:boundary-heads}

\begin{figure}[h]
\centering
\includegraphics[width=\linewidth]{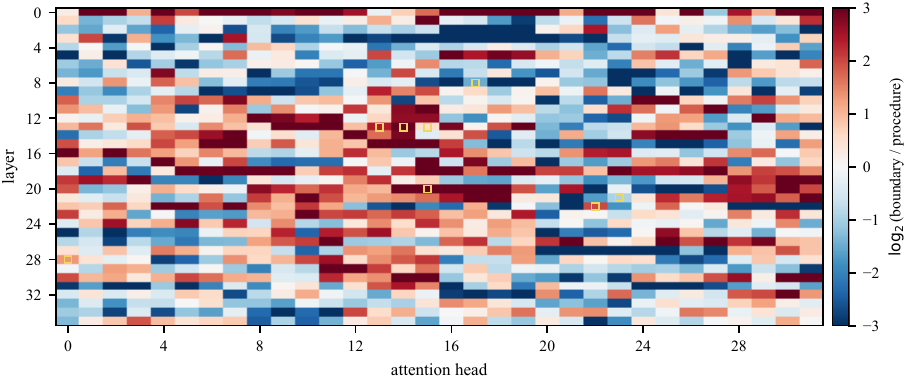}
\caption{\textbf{Head-level boundary attention.} Per-layer/per-head $\log_2$(boundary/procedure) attention for BASM in \Twrong{} at $k=2$. Boundary reading is concentrated in identifiable heads rather than uniformly spread across the prompt, consistent with a state-conditioned attention modulator.}
\label{fig:app-heatmap}
\end{figure}

\paragraph{Figure~\ref{fig:app-heatmap}: boundary reading is concentrated, not diffuse.}
Figure~\ref{fig:app-heatmap} plots the per-layer/per-head $\log_2$ boundary/procedure ratio for BASM in \Twrong{} at $k=2$. The map is not uniform. Several regions, especially in middle and later layers, assign substantially more decision-token attention to boundary spans than to procedure spans. This pattern is theoretically meaningful. If BASM were only improving behavior through global prompt length or a uniform caution prior, the attention map would be broadly smooth. Instead, the model appears to recruit a subset of attention heads that preferentially read the boundary when the retrieved procedure is semantically close but unsafe.

This head-level structure also explains why span-level knockout is effective. The boundary signal is not spread evenly across all tokens; it is routed through identifiable attention pathways that can be disrupted. These heads behave like a state-conditioned modulator: they do not erase the retrieved procedure, but they change how much the decision token should trust that procedure under the current state.

\begin{figure}[h]
\centering
\includegraphics[width=\linewidth]{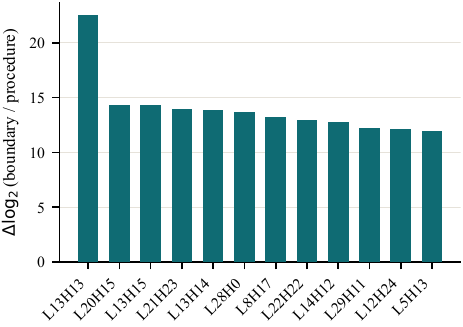}
\caption{\textbf{Candidate boundary heads.} Heads ranked by the increase in boundary/procedure log-ratio from \Tok{} to \Twrong{}. The largest shifts identify attention heads that become boundary-seeking exactly when procedural imitation becomes risky.}
\label{fig:app-boundary-heads}
\end{figure}

\paragraph{Figure~\ref{fig:app-boundary-heads}: candidate heads specialize in the risky-state transition.}
Figure~\ref{fig:app-boundary-heads} ranks heads by the increase in boundary/procedure log-ratio from \Tok{} to \Twrong{}. The strongest heads show large positive shifts, meaning they change behavior when the same kind of retrieved skill moves from a safe state to a risky one. This is a more precise signature than high boundary attention alone. A head that always attends to boundary text could simply be responding to formatting. A head whose boundary attention increases specifically from \Tok{} to \Twrong{} is instead sensitive to the state transition that defines the Skill Imitation Trap.

These candidate heads provide a concrete mechanistic locus for BASM's effect. The main causal claim remains at the span level, because span-level knockout directly tests the information channel used by the decision token. The head ranking goes one step deeper: it shows that the boundary channel is not an abstract metaphor, but an observable pattern in the model's attention computation. This strengthens the interpretation of BASM as a boundary-aware memory object rather than a longer prompt template.

\subsection{Boundary Mechanism, Behavior, and Control Comparisons}
\label{app:behavior-placebo}

\begin{figure}[h]
\centering
\includegraphics[width=\linewidth]{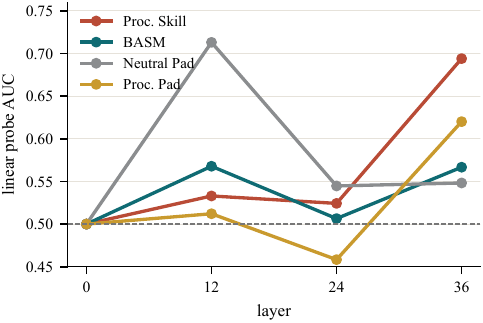}
\caption{\textbf{Residual-stream probe control.} Task-disjoint linear probes do not isolate a stable applicability vector. This supports the view that BASM works through dynamic decision-token attention to boundary text rather than through a static hidden-state label artifact.}
\label{fig:app-hidden-probe}
\end{figure}

\paragraph{Figure~\ref{fig:app-hidden-probe}: the effect is dynamic attention, not a static hidden-state label.}
Figure~\ref{fig:app-hidden-probe} reports a task-disjoint linear probe for skill applicability over residual-stream states. The probe does not reveal a stable, high-AUC applicability feature that is uniquely stronger under BASM. This result should not be read as a weakness of the method. Instead, it rules out a simpler and less interesting explanation: BASM is not merely injecting an easy linear label into the hidden state. The boundary signal is used dynamically at the decision token through attention to the retrieved skill text.

The Skill Imitation Trap occurs because retrieval gives the model a compelling procedure before the model has checked whether the current state satisfies the procedure's preconditions. BASM solves this by keeping the boundary available as contextual evidence and letting the decision token read it when the state demands it. A static residual-stream classifier is therefore not the expected mechanism. The negative linear-probe result is consistent with a stronger claim: boundary-aware skill memory changes the computation at the moment of action selection, rather than simply changing a global representation of the example.

\begin{figure}[h]
\centering
\includegraphics[width=\linewidth]{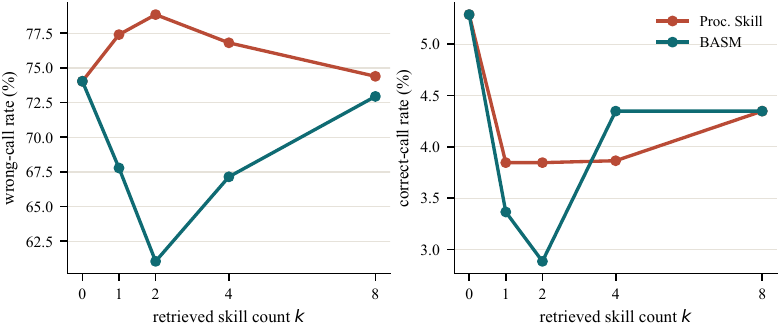}
\caption{\textbf{Recall-depth call rates.} The Skill Imitation Trap appears in realized candidate choices, not only in logits. Proc.\ Skill keeps wrong calls high near risky retrieval depths, whereas BASM suppresses those wrong calls by conditioning skill reuse on boundary evidence.}
\label{fig:app-call-rates}
\end{figure}

\paragraph{Figure~\ref{fig:app-call-rates}: the trap appears in realized candidate choices.}
Figure~\ref{fig:app-call-rates} verifies that the logit-margin story is not only a continuous-score artifact. As retrieval depth increases, Proc.\ Skill keeps the wrong-call rate high in \Twrong{}, showing the behavioral version of the Skill Imitation Trap: the more successful procedures the model sees, the more it is pulled toward calling a semantically related but invalid tool. BASM sharply lowers the wrong-call rate at the riskiest retrieval depths, especially around the depths where the logit probe shows the strongest trap.

The correct-call curve should be interpreted together with the wrong-call curve. In \Twrong{}, the desired behavior is often not simply ``call another tool''; it may be to avoid the retrieved tool, choose a control action, or wait for a repair path. Therefore, the key behavioral quantity is whether the model continues to execute the tempting wrong tool. By reducing wrong calls without relying on a generic no-tool bias, BASM demonstrates that boundary fields reshape the agent's action preference in the exact region where success-distilled memory is most dangerous.

\begin{figure}[h]
\centering
\includegraphics[width=\linewidth]{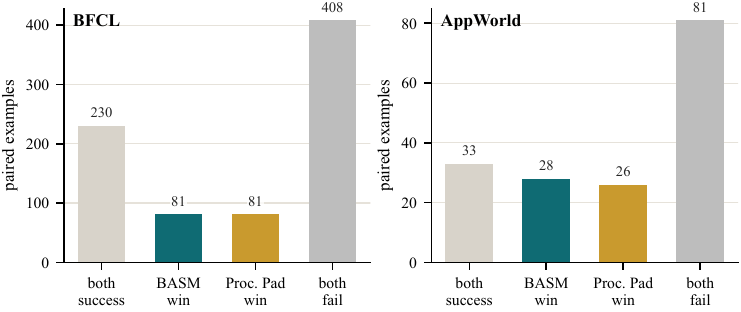}
\caption{\textbf{Equal-token control paired status.} Pairing against procedure padding isolates the value of boundary semantics under an equal context budget. BASM and padding trade wins on BFCL, while BASM is slightly ahead on AppWorld, consistent with the claim that boundary fields matter most when procedural memory must be state-conditioned.}
\label{fig:app-placebo-paired}
\end{figure}

\paragraph{Figure~\ref{fig:app-placebo-paired}: equal-token pairing isolates boundary semantics.}
Figure~\ref{fig:app-placebo-paired} compares BASM with a strong equal-token procedure-padding control at the paired-example level. This is a demanding control because procedure padding still contains task-relevant procedural text rather than empty filler. On BFCL, the paired wins are balanced between BASM and padding, while many examples are jointly solved or jointly failed. This observation does not undermine the mechanism result; instead, it clarifies the class of problems addressed by boundary information. BFCL contains many cases where extra procedural evidence can help with ordinary function selection, but the probe results show that when the retrieved procedure is plausible yet unsafe, boundary semantics is what suppresses the wrong-tool margin.

The AppWorld paired result is especially informative because the environment is stateful and multi-step. There, BASM wins more examples than procedure padding, matching the role of recovery notes and avoidance rules in long-horizon interaction. The paired analysis therefore supports a more precise conclusion: successful experience is valuable, but successful experience alone is incomplete. In states that require checking applicability, detecting risk cues, or repairing a local failure, a skill must carry its own boundary. BASM supplies that missing boundary, turning memory from a source of overconfident imitation into a source of controlled transfer.

\paragraph{Appendix takeaway.}
The appendix strengthens the main mechanism chain. Figure~\ref{fig:app-attention-ratio} shows that boundary text is read more in risky and repair states. Figure~\ref{fig:app-logit-margins} shows that this reading suppresses wrong-tool preference beyond equal-token controls. Figures~\ref{fig:app-knockout-8b} and~\ref{fig:app-knockout-14b} show that removing boundary attention pushes the model back toward imitation. Figures~\ref{fig:app-heatmap} and~\ref{fig:app-boundary-heads} show that the signal is routed through identifiable attention heads. Figure~\ref{fig:app-hidden-probe} rules out a static linear-feature explanation, and Figures~\ref{fig:app-call-rates}--\ref{fig:app-placebo-paired} connect the probe analysis to realized calls and paired outcomes. Taken together, these results support a precise conclusion: more memory is not automatically better. Memory becomes reliable only when a retrieved skill also tells the agent when not to imitate it.

\end{document}